\documentclass[10pt,twocolumn,letterpaper]{article}

\usepackage{cvpr}              

\usepackage{graphicx}
\usepackage{amsmath}
\usepackage{amssymb}
\usepackage{booktabs}

\usepackage{multirow}
\usepackage{algpseudocode}
\usepackage{algorithmicx,algorithm}
\usepackage[pagebackref,breaklinks,colorlinks]{hyperref}

\usepackage[capitalize]{cleveref}
\crefname{section}{Sec.}{Secs.}
\Crefname{section}{Section}{Sections}
\Crefname{table}{Table}{Tables}
\crefname{table}{Tab.}{Tabs.}

\def\cvprPaperID{6967} 
\def\confName{CVPR}
\def\confYear{2022}

\begin{document}

\title{GreedyFool: Multi-Factor Imperceptibility and Its Application to Designing a Black-box Adversarial Attack}

\author{Hui Liu \; Bo Zhao \; Minzhi Ji \\
School of Cyber Science and Engineering, Wuhan University\\
Wuhan, 430072 China\\
{\tt\small \{liuh824, zhaobo, jiminzhi\}@whu.edu.cn }
\and
Peng Liu\\
College of Information Sciences and Technology Pennsylvania State University\\
PA, 16801 US\\
{\tt\small pliu@ist.psu.edu}
}
\maketitle

\begin{abstract}

Adversarial examples are well-designed input samples, in which perturbations are imperceptible to the human eyes, but easily mislead the output of deep neural networks (DNNs). Existing works synthesize adversarial examples by leveraging simple metrics to penalize perturbations, that lack sufficient consideration of the human visual system (HVS), which produces noticeable artifacts. To explore why the perturbations are visible, this paper summarizes four primary factors affecting the perceptibility of human eyes. Based on this investigation, we design a multi-factor metric \emph{MulFactorLoss} for measuring the perceptual loss between benign examples and adversarial ones. In order to test the imperceptibility of the multi-factor metric, we propose a novel black-box adversarial attack that is referred to as GreedyFool. GreedyFool applies differential evolution to evaluate the effects of perturbed pixels on the confidence of a target DNN, and introduces greedy approximation to automatically generate adversarial perturbations. We conduct extensive experiments on the ImageNet and CIFRA-10 datasets and a comprehensive user study with 60 participants. The experimental results demonstrate that \emph{MulFactorLoss} is a more imperceptible metric than the existing pixelwise metrics, and GreedyFool achieves a 100\% success rate in a black-box manner.

\end{abstract}

\section{Introduction}
\label{sec:intro}
Existence of adversarial examples \cite{1, 2, 3, 4, 5, 6, 7, 8, qiu, laidlaw} is one of the most important findings in deep learning research \cite{9, 10, 11, 12, 13}. Since adversarial examples could enable researchers to gain profound understandings about the nature of deep learning models and how to safely deploy them in real-world AI systems, they have attracted a great amount of interests in the research community in recent years. Based on a widely-accepted notion of adversarial examples, in which certain changes to a sample of input data (e.g. an image) are (largely) {\bf imperceptible} to the human eye, but cause a deep learning model to {\bf misclassify} (or mis-predict) the input, researchers have been answering two basic research questions: (1) Why could certain slight changes trigger a deep learning model to misclassify? (2) Why could the changes needed to ``fool'' a deep learning model be imperceptible to the human eye? Besides answering the two basic research questions, researchers have also been developing integrated approaches to generating adversarial examples. By ``integrated'', we mean that the research findings in answering both questions are leveraged. For example, when a deep neural network (DNN) \cite{14, 15, 16, 17} is used to classify images, generating adversarial examples could be formalized as an optimization problem with imperceptibility (i.e. small perceptual loss) as a main constraint \cite{18, 19}.

Although the first research question has been extensively studied in the literature \cite{20, 21, 22, 23, 24}, the second question is less full investigation. In fact, recent studies attempted to study the second question from different angles \cite{ 3, qiu, laidlaw, 26, 29, zhao}. However, what are the primary factors affecting the perceptibility of human eyes? how do these factors combine to better measure the imperceptibility of perturbations? They does not give a comprehensive answer. In order to clearly show the limitations of the existing works, let's firstly summarize the primary factors affecting the HVS. To our best knowledge, the past (psychophysics) researches on the HVS have identified four primary factors affecting the {\bf perceptibility} of human eyes: (\emph{F1}) Just noticeable distortion (JND) \cite{25, 26}. Human eyes cannot sense a stimulus below the JND, which quantifies the sensitivity of the HVS to different background luminance. (\emph{F2}) Weber-Fechner law \cite{27, 28}. This law reveals important principle in psychophysics, which describes the logarithmic mapping between the magnitude of a physical stimulus and its perceived intensity. (\emph{F3}) Texture masking \cite{29}. Human eyes are more sensitive to perturbations on pixels in smooth regions than those in textured regions. (\emph{F4}) Channel modulation \cite{zhao, 30, 31}. Human vision has different sensitivity to the perturbations in different color channels.

\begin{itemize}
\item {\bf Key observation.} These four factors clearly show that the perceptibility of human eyes is a {\bf synthesized ability}; the perceptibility of human eyes is too complex to be reliably measured by a single metric.
\end{itemize}

Based on this key observation, we found that the limitations of the existing works are primarily due to the fact that they are all relying on a single metric. First, most of the existing works utilize distance metrics of $L_p$ norms ($L_0$, $L_2$ and $L_{\infty}$ norms) to measure imperceptibility. For example, Papernot et al. \cite{19} generated adversarial examples based on a precise understanding of the mapping between inputs and outputs of DNNs by $L_2$ norm. The fast gradient sign method (FGSM) \cite{20} computed one-step gradient to synthesize adversarial examples by $L_{\infty}$ norm. Carlini et al. \cite{8} used three distance metrics $L_0$, $L_2$ and $L_{\infty}$ to quantify imperceptibility. Regarding the limitations of these works, it is obvious that they do not treat perceptibility of the human eyes as a synthesized ability. As a result, the measured perceptual loss could be conflicting with factors \emph{F1}-\emph{F4} for some images. Second, Luo et al. \cite{29} defined perturbation sensitivity distance (PSD), taking the HVS into consideration, to measure perceptual loss between the benign examples and the adversarial ones. Zhao et al. \cite{zhao} minimized perturbation size with respect to perceptual color distance in RGB space to generate large yet imperceptible adversarial perturbations. Unfortunately, these works again does not treat perceptibility of human eyes as a synthesized ability. As a result, the measured perceptual loss in \cite{29} could be conflicting with factors \emph{F1}, \emph{F2}, and \emph{F4}. And in \cite{zhao}, the measured perceptual loss could not consider factors \emph{F1}-\emph{F3}.

To effectively address the limitations of the existing methods, this work seeks to integrate JND, Weber-Fechner law, texture masking and channel modulation and design a {\bf synthesized metric} to measure the perceptual loss caused by adversarial perturbations. To the best of our knowledge, this is the {\em first} metric ever proposed to measure perceptibility as a synthesized ability of human eyes in the context of generating adversarial examples. In order to design the synthesized metric, we first investigate how the four primary factors (JND, Weber-Fechner law, texture masking and channel modulation) are complementary to each other. We then design a pixelwise multi-factor metric for measuring perceptual loss caused by perturbations.

In order to see whether the synthesized metric can let the generated adversarial examples ``enjoy'' much improved imperceptibility, the existing works could be classified into two categories: back-propagation \cite{6, 7, 20, 21, 18} and forward-propagation \cite{5, 19, 29}. Back-propagation utilizes output error and gradient information to compute adversarial perturbations, eg. L-BFGS \cite{20}, FGSM \cite{21}, etc. This type of attack results in whole pixel perturbations, thus, it is not suitable for optimizing pixelwise metrics. Forward-propagation computes a direct mapping from the input to the output to achieve an explicit adversarial goal. Representative studies include Jacobian-based saliency map attack (JSMA) \cite{19}, evolutionary algorithms \cite{5}, etc. Since JSMA utilizes the forward derivative to construct adversarial saliency maps, it can not be deployed in a black-box scenario. Evolutionary algorithms are a set of heuristic search methods, among which differential evolution has been proved to be a reliable tool to generate pixelwise perturbations \cite{5}. However, if attackers fail to set the appropriate number of perturbed pixels, either the adversarial attack fails or the perturbations are unnecessarily large.

In order to overcome these limitations, assuming that the attacker does not have any inner information of the target DNNs, we apply a forward-propagation combining differential evolution and greedy approximation \cite{foolchecker} to minimize the proposed pixelwise multi-factor metric. Greedy approximation allows GreedyFool to automatically find which pixels to perturb and what magnitude to modify effectively, guaranteeing successful black-box attacks with less perceptual loss.

In summary, this paper makes the following contributions:
\begin{itemize}
\item We investigate the HVS and design a synthesized metric to calculate the perceptual loss between the benign examples and the adversarial ones. The synthesized metric can let the generated adversarial examples ``enjoy'' much improved imperceptibility than the existing methods.
\item To verify the advantages of the synthesized metric, We propose a pixelwise adversarial attack that is referred to as GreedyFool. GreedyFool attacks a target DNN only relying on its confidence. Thus, GreedyFool can attack more types of DNNs in a black-box manner. Since there are no budget constraints on the perceptual loss, GreedyFool achieves a 100\% success rate for both non-targeted attacks and targeted attacks. The code is available on: https://github.com/LiuHwell/greedyfool.git.
\end{itemize}

The remainder of this paper is organized as follows. We present the preliminaries in Sec.~\ref{sec:prel} and give the careful investigation into the HVS in Sec.~\ref{sec:hvs}. In Sec.~\ref{sec:method}, we give details about the design and the implementation of GreedyFool. The experimental evaluations are presented in Sec.~\ref{sec:evalu}. Finally, we conclude the paper in Sec.~\ref{sec:conclusion}. 

\section{Preliminaries}
\label{sec:prel}
\subsection{Deep neural network}
Deep neural networks \cite{14, 15, 16, 17} are usually constructed with multiple neural network layers. A neural network layer consists of a set of perceptrons and each perceptron maps a set of inputs to output values with an activation function, e.g., Sigmoid, ReLU. DNNs can be formed in a chain.
\begin{equation}
f(x,\theta)=f^{(k)}(\dots f^{(2)}(f^{(1)}(x,\theta_1),\theta_2),\theta_k)
\label{eq1}
\end{equation}
where $f^{(i)}(x,\theta_i)$ is the function of the $i^\prime$th layer of the network, where $i = 1, 2, ... , k$. \emph{x} is the input example and $\theta_i$ is the weight of the $i^\prime$th layer.

Convolutional neural networks is one of the most widely used neural networks, which comprise convolutional layers, pooling layers and fully connected layers. 
In order to verify the performance of GreedyFool, we perturb the CIFAR-10 dataset over DenseNet, and the ImageNet dataset \cite{imagenet} over Inception V3 \cite{inception}. 


\subsection{Adversarial attacks}
Adversarial attacks \cite{5, 6, 7, 18, 19} aim to mislead the DNN-based classifier to an incorrect label by adding small perturbations in the benign examples, even if the perturbations are barely recognizable by human eyes. Adversarial attacks would be formulated as a box-constrained optimization problem, that is,
\begin{equation}
\begin{aligned}
{\rm min}\quad&\rVert\delta\rVert_p \\
s.t.\quad&f(X)=l \\
&f(X^\prime)=l^\prime \\
&l\neq l^\prime \\
&X^\prime=X+\delta \in D\\
\end{aligned}
\label{eq2}
\end{equation}
where a trained DNN model \emph{f} predicts the benign example \emph{X} and the adversarial example $X^\prime$ into \emph{l} and $l^\prime$. $\rVert \cdot \rVert$ denotes the perceptual loss between two examples, and $\delta$ denotes the perturbation matrix that is added to the benign example for synthesizing the adversarial example, which remains in the benign domain $D$. 

\subsection{$L_p$ norms}
Definition of adversarial examples requires a distance metric to quantify similarity between the benign image and the adversarial image. Existing works mainly apply $L_p$ norms (e.g. $L_0$, $L_2$ and $L_{\infty}$) to measure the magnitude of the perturbation $\delta$ \cite{8, 19, 20}, that is,
\begin{equation}
\rVert\delta\rVert_{p}=\left(\sum_{i=1}^{n}\left|\delta_{i}\right|^{p}\right)^{\frac{1}{p}}
\label{eq3}
\end{equation}
where $L_0$ norm measures the number of pixels perturbed in an image. $L_2$ measures the Euclidean distance between the benign example and the adversarial example. $L_{\infty}$ norm denotes the maximum for all vector elements $\left|\delta_{i}\right|:\rVert\delta\rVert_{\infty}=max(\left|\delta_{i}\right|)$. In these definitions, the small $L_p$ norm value indicates the greater imperceptibility of perturbations.


\subsection{Differential evolution}
Differential evolution \cite{5, 32, 33} is a stochastic population-based algorithm for solving global optimization problems. The framework of differential evolution consists of the iteration of 3 operations: ``one-to-one-spawning" selection, mutation, and crossover. The ``one-to-one-spawning" selection mechanism replaces the parent solutions with the fitter child solutions. During each iteration, the algorithm mutates each candidate solution by mixing with other candidate solutions to create a trial candidate. There are several mutation strategies for creating trial candidates. Among these strategies, the ``rand/1" strategy is proved to be feasible in one-pixel attack \cite{5}, which is defined in Eq.~\ref{eq4}.
\begin{equation}
x_{i}(g+1)=x_{r1}(g)+F\left(x_{r2}(g)-x_{r3}(g)\right), r1 \neq r2 \neq r3
\label{eq4}
\end{equation}
where $r1$, $r2$, and $r3$ are random numbers and \emph{F} is the scale parameter, which is set to 0.5. Three candidate solutions $x_{r1}(g)$, $x_{r2}(g)$ and $x_{r3}(g)$ from the $g^\prime$th generation are randomly chosen to generate a new candidate solution $x_{i}(g+1)$ of the $g+1^\prime$th generation. Note that the crossover strategy is not included in our method as in \cite{5}.





\section{Human visual system}
\label{sec:hvs}
The human visual system is a multichannel model with characteristics of multifrequency channel decomposition \cite{gulina}. The sensitivity of human eyes to perturbations is affected by several factors. We investigate the HVS and summarize four primary factors that affecting human vision in the image domain: JND, Weber-Fechner law, texture masking and channel modulation.

\subsection{Just noticeable distortion}
\begin{figure}[htb]
\centering
\includegraphics[width=0.75\columnwidth]{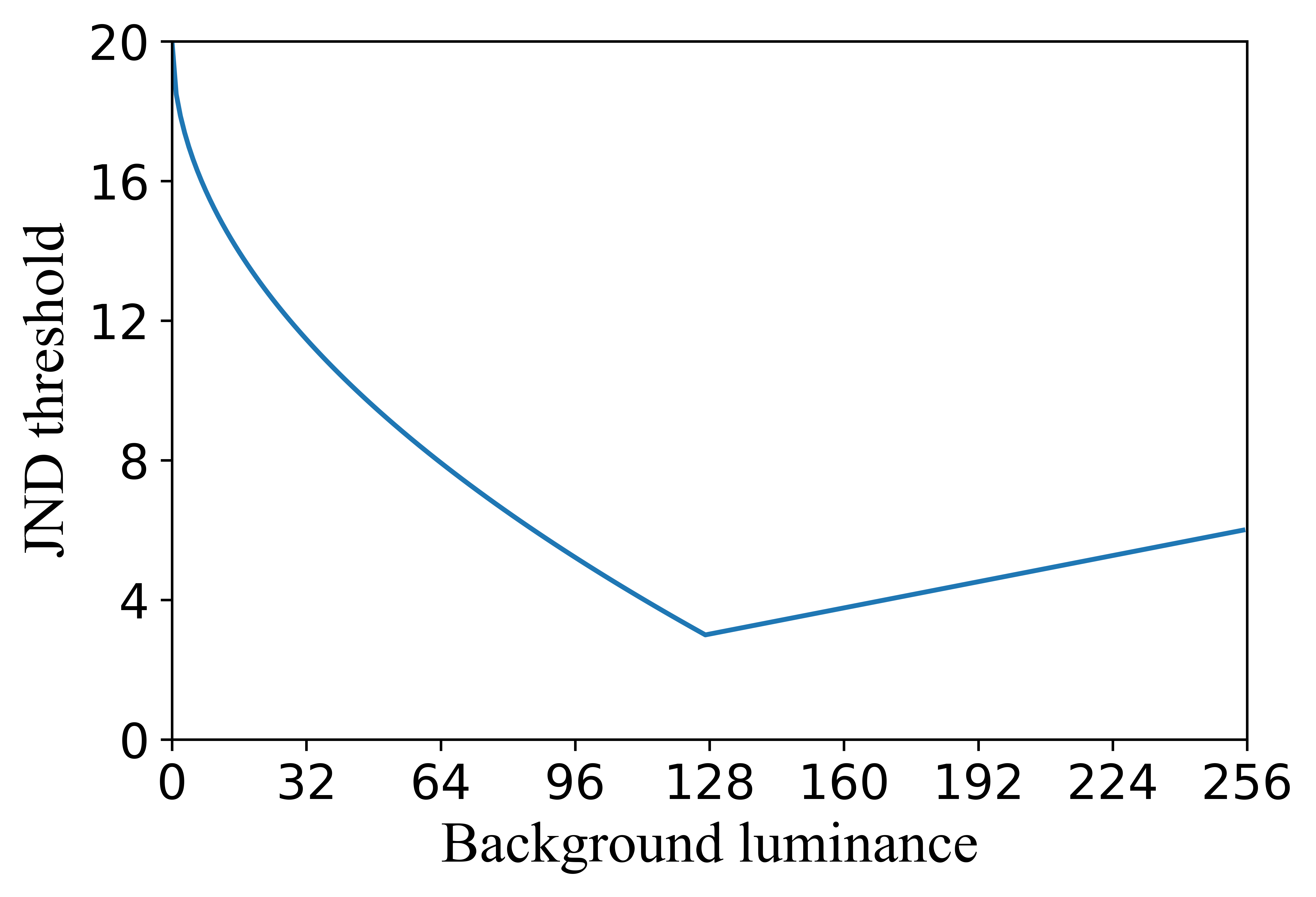}
\caption{Illustration of the JND. When the maximum luminance is 127, the JND threshold reaches a minimum 3.}
\label{fig3}
\end{figure}

Human eyes cannot perceive a stimulus below the just noticeable distortion \cite{25, 26} threshold around a pixel in images.  
A larger value of the JND threshold indicates that more noise can be hidden. 
The visibility threshold of the JND is formulated in Eq.~\ref{eq5}, and shown as the curve in Fig.~\ref{fig3}.
\begin{equation}
jnd(x, y)=
\begin{cases}
17(1-\sqrt{\frac{\bar{I}_{n \times n}(x, y)}{127}})+3, \mbox{if }\bar{I}_{n \times n}(x, y) \leq 127 \\
\frac{3}{128}\left(\bar{I}_{n \times n}(x, y)-127\right)+3, \mbox{otherwise}
\end{cases}
\label{eq5}
\end{equation}
where $ \bar{I}_{n \times n}(x, y) $ denotes the maximum luminance among a $n \times n$ window at the coordinates $(x, y)$. Especially, we set $n=3$ and calculate the maximum luminance of the $(x, y)$ pixel and its 8 neighbors.

Figure~\ref{fig3} reveals the nonlinear relationship between the JND threshold and the maximum background luminance. Even if the magnitude of perturbations is the same, there are significant differences in the perceived intensity of human eyes in different luminance backgrounds. 

\subsection{Weber-Fechner law}
Weber-Fechner law \cite{27, 28} has been proposed in the field of psychophysics to quantify relationships between any stimulus and the perceived response by individuals. The Weber law asserts that the just noticeable stimulus difference $\Delta I$ maintains a constant ratio with respect to the intensity of the comparison stimulus \emph{I}. 
On the assumption that the difference threshold represents a unit change in sensation $\Delta S$, Fechner defined Weber's law as:
\begin{equation}
\Delta S=k \frac{\Delta I}{I}
\label{eq6}
\end{equation}

Integrating this formula, a logarithmic relation called the psychophysical law is generated, as shown in Eq.~\ref{eq7}.
\begin{equation}
S=k\ln I+C
\label{eq7}
\end{equation}
where \emph{k} and \emph{C} are hyperparameters, \emph{I} denotes the magnitude of a physical stimulus and \emph{S} denotes the corresponding perceived intensity. The Weber-Fechner law revealed the logarithmic mapping between the magnitude of a physical stimulus and its perceived intensity, rather than the absolute magnitude value as demonstrated in \cite{29}.

\subsection{Texture masking}
\begin{figure}[!t]
\centering

  \begin{subfigure}{0.32\linewidth}
    \includegraphics[width=.9\linewidth]{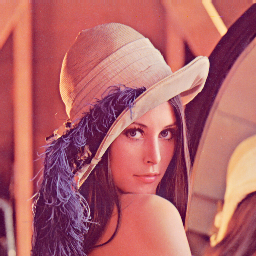}
    \label{fig4-a}
    \caption{}
  \end{subfigure}
  \begin{subfigure}{0.32\linewidth}
    \includegraphics[width=.9\linewidth]{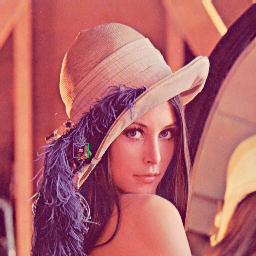}
    \label{fig4-b}
    \caption{}
  \end{subfigure}
  \begin{subfigure}{0.32\linewidth}
    \includegraphics[width=.9\linewidth]{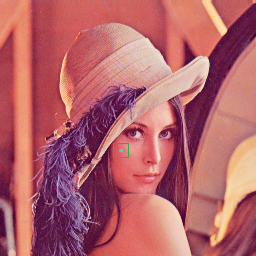}
    \label{fig4-c}
    \caption{}
  \end{subfigure}

  \caption{Perceptual loss of the perturbations with the same magnitude in different regions. The green line box marks perturbations in perturbed images. (a) benign example, (b) perturbed image at textured regions, (c) perturbed image at smooth regions.}
  \label{fig4}
\end{figure}

According to texture masking theory \cite{29}, human eyes are more sensitive to perturbations on pixels in smooth regions than those in textured regions. We add 3$\times$3 window perturbations with 100 magnitudes in the smooth region and the textured region for comparison of the perceptual loss. Figure~\ref{fig4} plots the benign example, the perturbed image at textured regions and perturbed image at smooth regions from left to right. When perturbations with the same magnitude are added to the image, the perturbations in smooth regions are easily detected by human eyes, while those in textured regions are difficult to recognize.

The standard deviation is a commonly employed quantity for measuring the texture masking of an image, which is proven to be effective in evaluating the perceptual loss of adversarial examples. The paper computes the standard deviation of a pixel $p_i$ in a $n\times n$ window as shown in Eq.~\ref{eq8}.
\begin{equation}
SD\left(p_{i}\right)=\sqrt{\frac{\sum_{p_{i} \in S_{i}}\left(p_{i}-\mu\right)^{2}}{n^{2}}}
\label{eq8}
\end{equation}
where $S_i$ is the set of pixels in the $n\times n$ window, and $\mu$ comprise the average values of pixels in the region. The standard deviation $SD(p_{i})$ is the variance of pixel $p_{i}$ among $S_{i}$. In this paper, we set $n=3$ and calculate the standard deviation of pixel $p_{i}$ and its 8 neighbors. 

\subsection{Channel modulation}
\begin{figure}[!t]
\centering
  \begin{subfigure}{0.36\linewidth}
    \includegraphics[width=.9\linewidth]{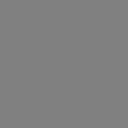}
    \label{fig5-a}
    \caption{Benign example}
  \end{subfigure}
  \hspace{0.6cm}
  \begin{subfigure}{0.36\linewidth}
    \includegraphics[width=.9\linewidth]{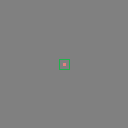}
    \label{fig5-b}
    \caption{Red channel}
  \end{subfigure}

  \begin{subfigure}{0.36\linewidth}
    \includegraphics[width=.9\linewidth]{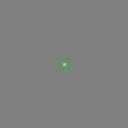}
    \label{fig5-c}
    \caption{Green channel}
  \end{subfigure}
  \hspace{0.6cm}
  \begin{subfigure}{0.36\linewidth}
    \includegraphics[width=.9\linewidth]{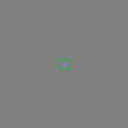}
    \label{fig5-d}
    \caption{Blue channel}
  \end{subfigure}
\caption{Perceptual loss of the perturbations with the same magnitude in different channels. The green line box marks perturbations in perturbed examples.}
\label{fig5}
\end{figure}

According to spectral sensitivity theory \cite{30, 31}, human eyes consist of red, green and blue cone cells, and the cone is sensitized to different ranges of wavelengths to provide a range of color perception. The human eyes are the most sensitive to green, followed by red, and the least sensitive to blue. We conduct perturbations with the same magnitude in three color channels. As shown in Fig~\ref{fig5}, the test proves that there are differences in the sensitivity of human eyes to color channels.

We introduce channel modulation to quantify the weight of perturbations in three color channels. Channel modulation refers to a constrained linear combination of red, green and blue channels based on decolorization theory \cite{31}, which is formulated in Eq.~\ref{eq9}.
\begin{equation}
\begin{aligned}
\varpi&=\lambda_{r} I_{r}+\lambda_{g} I_{g}+\lambda_{b} I_{b} \\
s.t.&\quad\lambda_{r}+\lambda_{g}+\lambda_{b}=1 \\
~&\quad\lambda_{r} \geq 0, \lambda_{g} \geq 0, \lambda_{b} \geq 0
\end{aligned}
\label{eq9}
\end{equation}
where $I_{r}$, $I_{g}$ and $I_{b}$ are the red channel, green channel and blue channel respectively, and $\varpi$ is the result of the channel modulation. The non-negative numbers $\lambda_{r}$, $\lambda_{g}$ and $\lambda_{b}$ are channel weights that sum to 1. In the classical RGB2GRAY conversion model \cite{31}, the weights are fixed as ${\lambda_{r}=0.299, \lambda_{g}=0.587, \lambda_{b}=0.114}$.

\section{Methodology}
\label{sec:method}
Adversarial attacks could be formulated as an optimization problem with constraints. This problem involves the definition of the perceptual loss and the solution for the optimal problem. In this section, we describe a definition of the multi-factor perceptual loss and present a forward-propagation that combines differential evolution with greedy approximation
\subsection{Multi-factor perceptual loss}
In the investigation into the HVS, we discover that the perceptual loss is primarily affected by four factors: JND, Weber-Fechner law, texture masking and channel modulation. 
In computer vision, the application of Weber-Fechner law needs to consider the difference in the perceived intensity of the human eyes with different luminance backgrounds. The visibility of the human eyes to perturbations of images has a positive correlation with a stimulus and a negative correlation with JND. Thus, we combine Eqs.~\ref{eq5} and ~\ref{eq6} to redefine this correlation, as shown in Eq.~\ref{eq10}.
\begin{equation}
\Delta Ps=k \frac{\Delta I}{JND(I)}
\label{eq10}
\end{equation}
where \emph{k} is a constant according to Weber-Fechner law, $\Delta I$ is the just noticeable stimulus difference and $\Delta Ps$ is the corresponding change in visual sensation. Since $JND(I)$ is a discrete piecewise function, we sum over Eq.~\ref{eq10} rather than integrate, as shown in Eq.~\ref{eq11}.

\begin{equation}
Ps(I)=k \sum_{i=I_{0}}^{I-1} \frac{1}{JND(i)}+C
\label{eq11}
\end{equation}
where $Ps(I)$ is the perceptual stimulus, $I_0$ is the background luminance, and \emph{k} and \emph{C} are constants.

To construct the mapping between a physical stimulus and its perceptual stimulus, we define the perceptual stimulus in the interval [0, 255]. Obviously, a perturbation of magnitude 0 does not cause any perceptual perception to the human eyes and a perturbation of magnitude 255 could cause the most noticeable perceptual perception to the human eyes. Thus, the following equation could be constructed to compute the constants \emph{k} and \emph{C}, that is,
\begin{equation}
\left\{\begin{array}{l}
Ps(0)=C=0, \mbox{if } I=0 \\
Ps(255)=k\sum_{i=0}^{255}\frac{1}{JND(i)}+C=255, \mbox{if } I=255
\end{array}\right.
\label{eq12}
\end{equation}

By solving Eq.~\ref{eq12}, we obtain the parameters $C=0$ and $k=\frac{256}{\sum_{i=0}^{255} \frac{1}{JND(i)}}$.

Texture masking reflects that perturbations in the region with high standard deviation are more imperceptible than those in the region with low standard deviation. The visibility of human eyes to perturbations has a negative correlation with the standard deviation of the region in an image. Furthermore, considering channel modulation, 
we propose a synthesized metric for evaluating the perceptual loss as follows.
\begin{equation}
IntegLoss(p_{i})=\sum_{c \in(r, g, b)} \lambda_{c} \frac{Ps(p_{i})}{SD_{c}(p_{i})}
\label{eq13}
\end{equation}

Adversarial attacks usually add multiple pixel perturbations for the success rate. The perceptual loss between the benign image and the adversarial image is the sum of all pixelwise perceptual losses. Therefore, we sum all pixel-wise perceptual losses as follows.
\begin{equation}
MulFactorLoss(X, X^{\prime})=\sum_{i=1}^{N} \sum_{c \in(r, g, b)} \lambda_{c} \frac{Ps_{c}(p_{i})}{SD_{c}(p_{i})}
\label{eq14}
\end{equation}
where \emph{N} is the number of perturbed pixels in a benign image. $MulFactorLoss(X, X^{\prime})$ denotes the perceptual loss that integrates \emph{JND}, Weber-Fechner law, texture masking and channel modulation between the benign example \emph{X} and the adversarial example $X^{\prime}$. The small $MulFactorLoss$ value indicates the high perceptual similarity between the benign example and the adversarial example.
\subsection{Pixelwise objective function}
State-of-the-art adversarial attacks should allow DNNs to give the wrong output with a high confidence score by adding as few perturbations as possible. Therefore, we should choose pixels that can reduce the confidence of DNNs in the true class or increase that in the target class with the less perceptual loss. The pixelwise objective function, which is referred to as the perturbation priority, is defined to estimate the effect of perturbing a pixel as follows.
\begin{equation}
PertPriority\left(p_{i}\right)=\zeta \frac{P_{t}(X)-P_{t}\left(X^{\prime}\right)}{MulFactorLoss\left(p_{i}\right)}
\label{eq15}
\end{equation}
where $\zeta$ is a control parameter. $P_{t}$ denotes the probability that the example belongs to the label \emph{t}. The adversarial example $X^{\prime}$ is synthesized by changing a pixel $p_{i}$ of benign example \emph{X}. When $t=l$, $\zeta=1$ and non-targeted attacks are executed, otherwise $\zeta=-1$ and targeted attacks are executed. The perturbation priority quantifies the effects of the current pixel $p_{i}$ perturbation on the confidence of the DNN-based classifier in the target class.

As can be seen from Eq.~\ref{eq15}, this objective function only relies on the DNN's confidence on the test image. Thus, GreedyFool can attack more types of DNNs in a black-box manner.
\subsection{Implementation}
To determine the pixels with high priority, the adversary has to choose which pixels to modify and what magnitudes to add. 
We encode the pixelwise perturbation into an array as a candidate solution $(x, y, r, g, b)$, which contains five elements: $x-y$ coordinates and RGB value of the perturbation. A brute-force approach has to search all dimensions and pixel values for the optimal value, which could take a prohibitively long time.

To reduce the search time, we introduce differential evolution to solve the optimal pixelwise objective function. GreedyFool sets the total population size to 200 and the number of generations to 60, resulting in 12,000 candidate solutions with perturbation priority. Greedy approximation is utilized to automatically obtain a set of perturbed pixels and synthesize the imperceptible adversarial example. The implementation of GreedyFool is presented as follows.

Step 1. Encoding the perturbation into candidate solution $(x, y, r, g, b)$, and randomly initializing 200 candidate solutions;

Step 2. Executing differential evolution during 60 generations to calculate the priority of candidate solutions;

Step 3. Greedy approximation is executed to choose a set of candidate solutions with the highest priority as adversarial perturbations. These perturbations are added to the benign image to synthesize its adversarial example.

\section{Experimental evaluation}
\label{sec:evalu}

\subsection{Experimental setup}
In this experiment, we implement GreedyFool in Python and conduct the adversarial attack in ImageNet over Inception V3, and CIFAR-10 over DenseNet. Due to the size 224$\times$224 of the preprocessed ImageNet and 32$\times$32 of CIFAR-10, and the pixel value interval [0, 255], the population of the differential evolution is initialized by uniform distributions $ U(1,224)$ for ImageNet and $ U(1,32)$ for CIFAR-10 to generate the $x-y$ coordinates and Gaussian distributions $N(\mu=128, \sigma=127)$ for pixel RGB values. The fitness function of the differential evolution is set to a pixelwise objective function in Eq.~\ref{eq15}. 
GreedyFool runs on an Intel Core I5 CPU 2.30 GHz, NVIDIA GeForce GTX 1060 and 8.0 GB of RAM computer that run on Windows 10 and Spyder (Python 3.6). 

\subsection{Adversarial examples on ImageNet}
\begin{figure*}[!t]
\centering

  \begin{subfigure}{0.19\linewidth}
    \includegraphics[width=.9\linewidth]{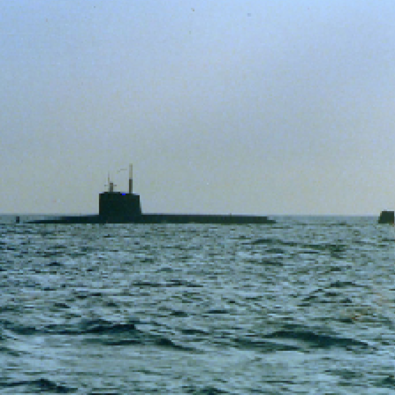}
    \label{fig6_1-a}
    \caption{}
  \end{subfigure}
  \begin{subfigure}{0.19\linewidth}
    \includegraphics[width=.9\linewidth]{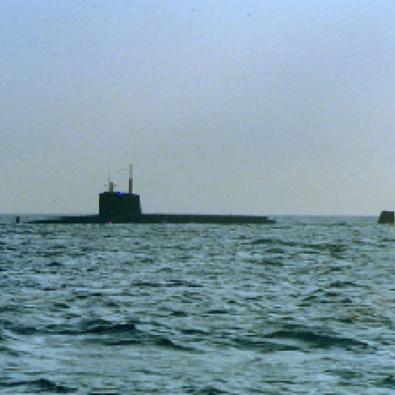}
    \label{fig6_1-b}
    \caption{}
  \end{subfigure}
  \begin{subfigure}{0.19\linewidth}
    \includegraphics[width=.9\linewidth]{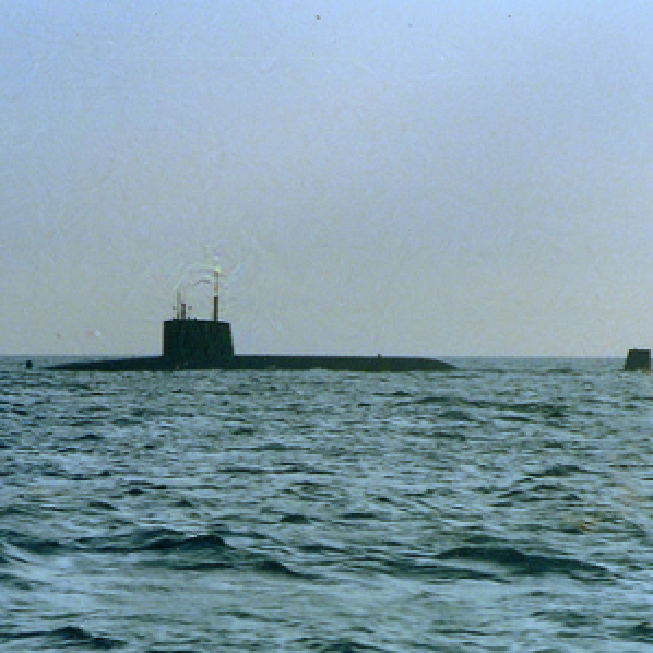}
    \label{fig6_1-c}
    \caption{}
  \end{subfigure}
  \begin{subfigure}{0.19\linewidth}
    \includegraphics[width=.9\linewidth]{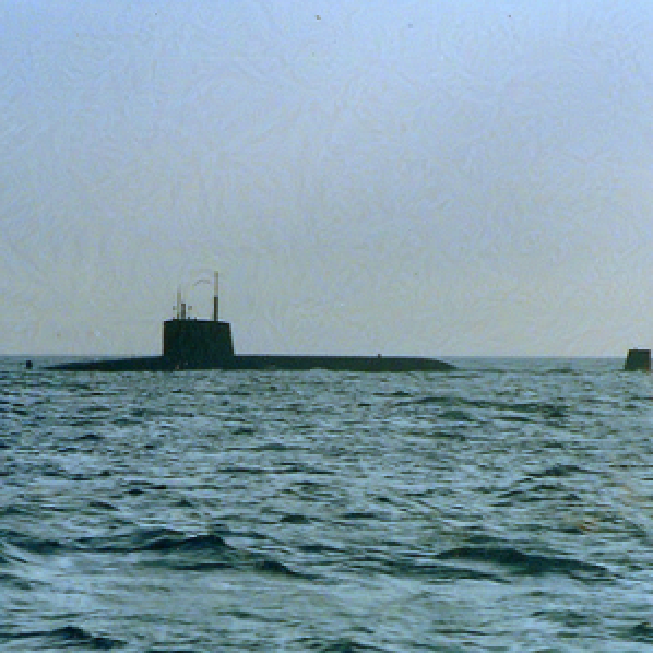}
    \label{fig6_1-d}
    \caption{}
  \end{subfigure}
  \begin{subfigure}{0.19\linewidth}
    \includegraphics[width=.9\linewidth]{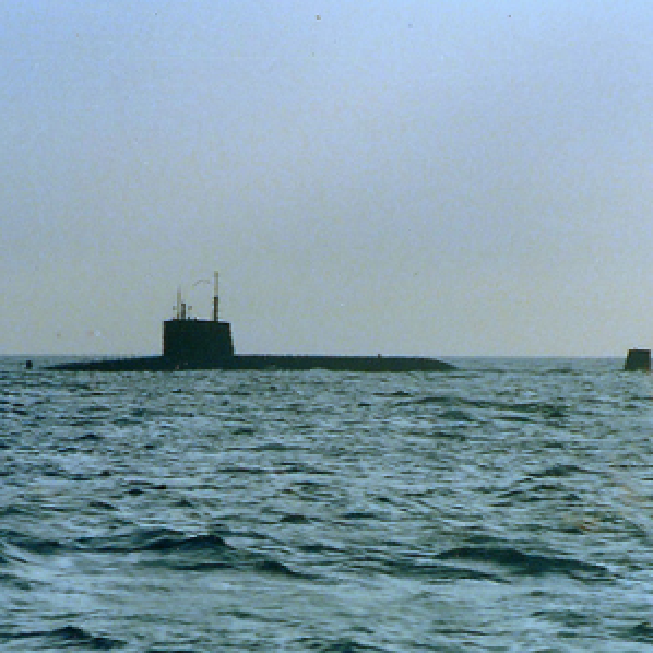}
    \label{fig6_1-e}
    \caption{}
  \end{subfigure}
\caption{Comparison of adversarial examples generated by GreedyFool, L-BFGS, FGSM, and Color-and-Edge-Aware on the same target label breakwater.}
\label{fig6_1}
\end{figure*}

We evaluate GreedyFool and compare it with 3 state-of-the-art adversarial attacks (L-BFGS \cite{20}, FGSM \cite{21}, Color-and-Edge-Aware \cite{bassett}) on the Inception V3 model for ImageNet classification task. Figure~\ref{fig6_1} shows the comparison of adversarial examples.
Close inspection reveals perturbation artefacts in the sky near the water's surface for all methods except GreedyFool. Table ~\ref{table1} lists \emph{MulFactorLoss} values of adversarial examples in Fig.~\ref{fig6_1}. The adversarial example generated by GreedyFool presents less visibility and \emph{MulFactorLoss} value than those generated by the other 3 attacks.

\begin{table}
\centering

\begin{tabular}{@{}lcccc@{}}
\toprule
Method & GreedyFool & \cite{20} & \cite{21} & \cite{bassett} \\
\midrule
\emph{MulFactorLoss} & 49.81 & 432.16 & 365.87 & 137.35 \\
\bottomrule
\end{tabular}

\caption{\emph{MulFactorLoss} values of adversarial examples on imageNet}
\label{table1}
\end{table}

\subsection{Human eye evaluation on CIFAR-10}
\begin{figure*}[htbp]
\centering

  \begin{subfigure}{0.13\linewidth}
    Benign images
  \end{subfigure}
  \begin{subfigure}{0.08\linewidth}
    \includegraphics[width=.9\linewidth]{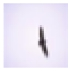}
  \end{subfigure}
  \begin{subfigure}{0.08\linewidth}
    \includegraphics[width=.9\linewidth]{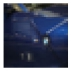}
  \end{subfigure}
  \begin{subfigure}{0.08\linewidth}
    \includegraphics[width=.9\linewidth]{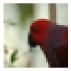}
  \end{subfigure}
  \begin{subfigure}{0.08\linewidth}
    \includegraphics[width=.9\linewidth]{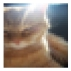}
  \end{subfigure}
  \begin{subfigure}{0.08\linewidth}
    \includegraphics[width=.9\linewidth]{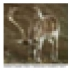}
  \end{subfigure}
  \begin{subfigure}{0.08\linewidth}
    \includegraphics[width=.9\linewidth]{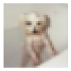}
  \end{subfigure}
  \begin{subfigure}{0.08\linewidth}
    \includegraphics[width=.9\linewidth]{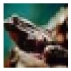}
  \end{subfigure}
  \begin{subfigure}{0.08\linewidth}
    \includegraphics[width=.9\linewidth]{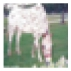}
  \end{subfigure}
  \begin{subfigure}{0.08\linewidth}
    \includegraphics[width=.9\linewidth]{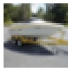}
  \end{subfigure}
  \begin{subfigure}{0.08\linewidth}
    \includegraphics[width=.9\linewidth]{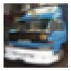}
  \end{subfigure}

  \begin{subfigure}{0.13\linewidth}
    GreedyFool
  \end{subfigure}
  \begin{subfigure}{0.08\linewidth}
    \includegraphics[width=.9\linewidth]{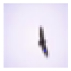}
  \end{subfigure}
  \begin{subfigure}{0.08\linewidth}
    \includegraphics[width=.9\linewidth]{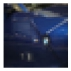}
  \end{subfigure}
  \begin{subfigure}{0.08\linewidth}
    \includegraphics[width=.9\linewidth]{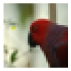}
  \end{subfigure}
  \begin{subfigure}{0.08\linewidth}
    \includegraphics[width=.9\linewidth]{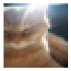}
  \end{subfigure}
  \begin{subfigure}{0.08\linewidth}
    \includegraphics[width=.9\linewidth]{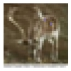}
  \end{subfigure}
  \begin{subfigure}{0.08\linewidth}
    \includegraphics[width=.9\linewidth]{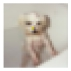}
  \end{subfigure}
  \begin{subfigure}{0.08\linewidth}
    \includegraphics[width=.9\linewidth]{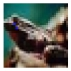}
  \end{subfigure}
  \begin{subfigure}{0.08\linewidth}
    \includegraphics[width=.9\linewidth]{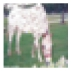}
  \end{subfigure}
  \begin{subfigure}{0.08\linewidth}
    \includegraphics[width=.9\linewidth]{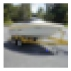}
  \end{subfigure}
  \begin{subfigure}{0.08\linewidth}
    \includegraphics[width=.9\linewidth]{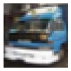}
  \end{subfigure}

  \begin{subfigure}{0.13\linewidth}
    One-pixel
  \end{subfigure}
  \begin{subfigure}{0.08\linewidth}
    \includegraphics[width=.9\linewidth]{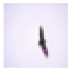}
  \end{subfigure}
  \begin{subfigure}{0.08\linewidth}
    \includegraphics[width=.9\linewidth]{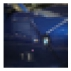}
  \end{subfigure}
  \begin{subfigure}{0.08\linewidth}
    \includegraphics[width=.9\linewidth]{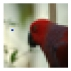}
  \end{subfigure}
  \begin{subfigure}{0.08\linewidth}
    \includegraphics[width=.9\linewidth]{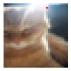}
  \end{subfigure}
  \begin{subfigure}{0.08\linewidth}
    \includegraphics[width=.9\linewidth]{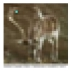}
  \end{subfigure}
  \begin{subfigure}{0.08\linewidth}
    \includegraphics[width=.9\linewidth]{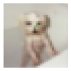}
  \end{subfigure}
  \begin{subfigure}{0.08\linewidth}
    \includegraphics[width=.9\linewidth]{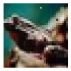}
  \end{subfigure}
  \begin{subfigure}{0.08\linewidth}
    \includegraphics[width=.9\linewidth]{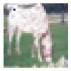}
  \end{subfigure}
  \begin{subfigure}{0.08\linewidth}
    \includegraphics[width=.9\linewidth]{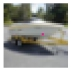}
  \end{subfigure}
  \begin{subfigure}{0.08\linewidth}
    \includegraphics[width=.9\linewidth]{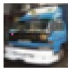}
  \end{subfigure}

  \begin{subfigure}{0.13\linewidth}
    PSD
  \end{subfigure}
  \begin{subfigure}{0.08\linewidth}
    \includegraphics[width=.9\linewidth]{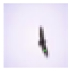}
  \end{subfigure}
  \begin{subfigure}{0.08\linewidth}
    \includegraphics[width=.9\linewidth]{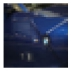}
  \end{subfigure}
  \begin{subfigure}{0.08\linewidth}
    \includegraphics[width=.9\linewidth]{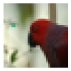}
  \end{subfigure}
  \begin{subfigure}{0.08\linewidth}
    \includegraphics[width=.9\linewidth]{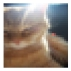}
  \end{subfigure}
  \begin{subfigure}{0.08\linewidth}
    \includegraphics[width=.9\linewidth]{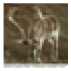}
  \end{subfigure}
  \begin{subfigure}{0.08\linewidth}
    \includegraphics[width=.9\linewidth]{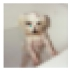}
  \end{subfigure}
  \begin{subfigure}{0.08\linewidth}
    \includegraphics[width=.9\linewidth]{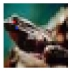}
  \end{subfigure}
  \begin{subfigure}{0.08\linewidth}
    \includegraphics[width=.9\linewidth]{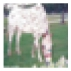}
  \end{subfigure}
  \begin{subfigure}{0.08\linewidth}
    \includegraphics[width=.9\linewidth]{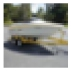}
  \end{subfigure}
  \begin{subfigure}{0.08\linewidth}
    \includegraphics[width=.9\linewidth]{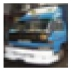}
  \end{subfigure}

\caption{Adversarial images synthesized by different attack methods against DenseNet in CIFAR-10. Adversarial examples in the second row crafted by our method are much more imperceptible than other examples from the following rows.}
\label{fig7}
\end{figure*}

To clearly present the details of the pixelwise perturbations, we generated adversarial examples on the CIFAR-10 dataset and carry out human eye tests. Similar to GreedyFool, both One-pixel attack \cite{5} and PSD attack \cite{29} belong to black-box pixelwise attack that use forward-propagation. Therefore, all three attacks are carried out on the same, random set of images to compare the imperceptibility of adversarial perturbations. We present 10 classes of images in Fig.~\ref{fig7}, which include benign images and their adversarial images against DenseNet. 

\begin{figure}[htb]
\centering
\includegraphics[width=0.75\columnwidth]{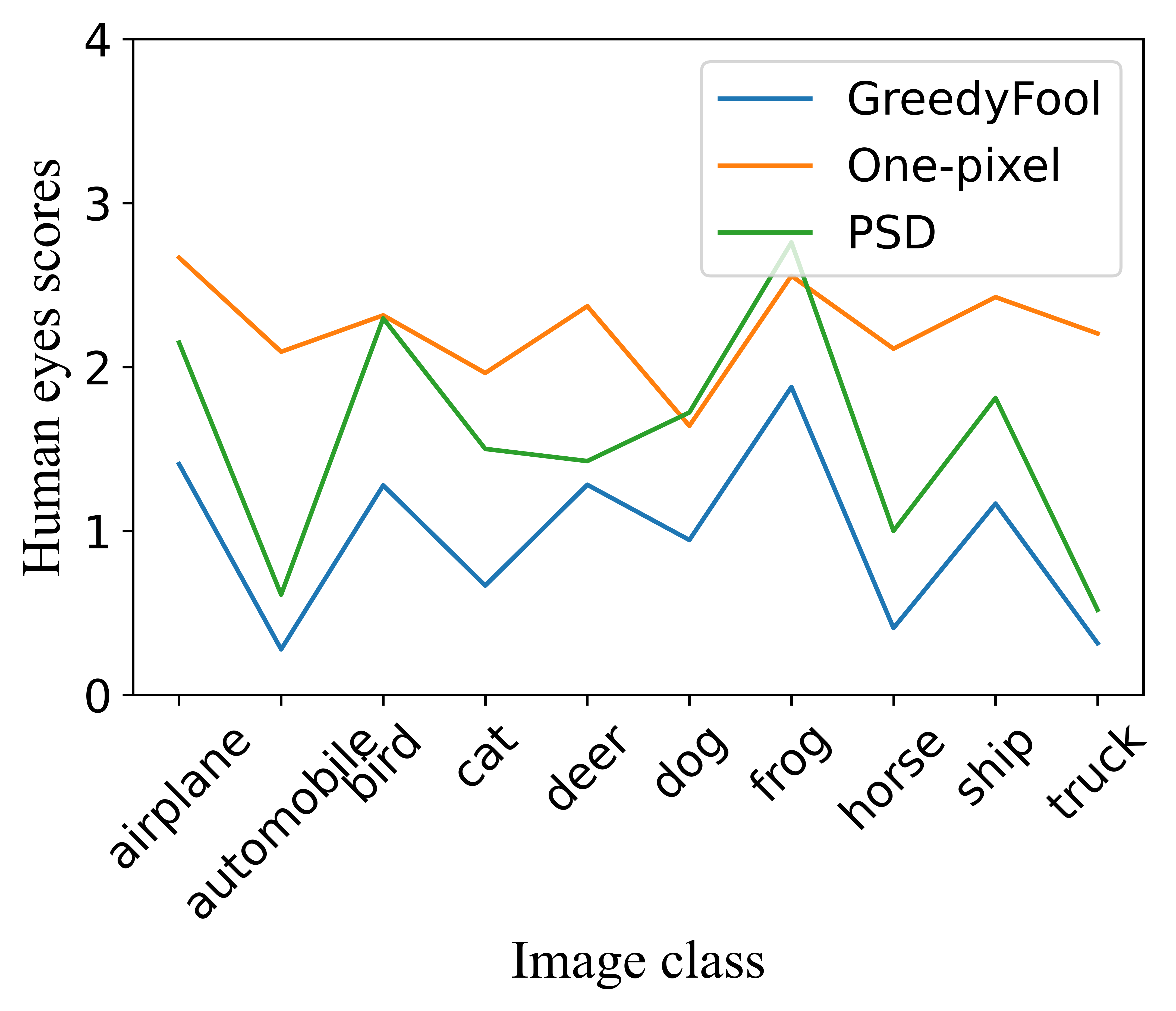}
\caption{Average scores comparison from human eyes evaluation between our attacks with existing attacks on DenseNet.}
\label{fig9}
\end{figure}

Imperceptibility is a subjective feeling that varies among individuals. To evaluate the visibility of adversarial perturbations, we design a scoring system based on the human eye evaluation and conduct an extensive user study. In this test, we generate 150 images for each attack over DenseNet, which are from 15 benign images targeting for 10 classes. We recruit 60 participants to score the similarity between benign images and their adversarial examples generated randomly by one of the aforementioned attacks. For each trial, the benign images are shown on a screen at a fixed size and the order of adversarial images is random. The participants are required to give a score from 0 to 10 in 5 seconds. A higher score denotes more noticeable adversarial perturbations. A score of 0 means no difference, while a score of 10 means a complete difference. We calculate the average score for each class of images from the human eye evaluation and plot them in Fig.~\ref{fig9} for comparison.

As shown in Fig.~\ref{fig9}, GreedyFool has the lowest score in any classes. One-pixel attack has the highest score in most classes, because human eyes can easily detect perturbed pixels. The results show that adversarial examples synthesized by GreedyFool have higher perceptual similarity to their benign images. It indicates that GreedyFool generates more imperceptible adversarial examples than other pixelwise methods.

Among the three types of attacks, One-pixel attack hardly considers the HVS. PSD attack considers the intensity and texture masking of benign images. GreedyFool combines more abundant and effective factors to construct an objective function of adversarial attacks, which fully considers the HVS. The human eye experiment demonstrates that the imperceptibility of adversarial examples could be further improved by combining multiple HVS metrics.

\subsection{Perceptual loss results}
A good perceptual loss should reliably reflect the perceptual similarity between the benign example and the adversarial example. Table~\ref{table2} lists \emph{MulFactorLoss} results of the adversarial examples generated by these three methods. The results show that GreedyFool could generate adversarial examples with smaller \emph{MulFactorLoss} values than PSD attack and One-pixel attack, which are consistent with the scores obtained from the human eye evaluation.

\begin{table}[htb]
\centering
\begin{tabular}{@{}lccc@{}}
\toprule
Method & \cite{5} & \cite{29} & GreedyFool \\
\midrule
airplane & 2.98 & 1.32 & 0.34 \\
automobile & 0.01 & 24.64 & 0.01 \\
bird & 11.09 & 3.45 & 2.41 \\
cat & 7.88 & 2.48 & 0.65 \\
deer & 7.32 & 0.67 & 0.40 \\
dog & 4.04 & 0.53 & 1.01 \\
frog & 10.87 & 2.84 & 1.55 \\
horse & 7.06 & 0.28 & 0.15 \\
ship & 13.98 & 0.17 & 0.04 \\
truck & 7.19 & 0.17 & 0.06 \\
\bottomrule
\end{tabular}
\caption{Multi-factor perceptual loss results of adversarial examples over DenseNet}
\label{table2}
\end{table}

Table~\ref{table3} lists the perceptual loss results of the adversarial examples generated by GreedyFool in Fig.~\ref{fig7} The results show that $L_p$ norms cannot always reliably quantify the imperceptibility of adversarial perturbations. The multi-factor perceptual loss \emph{MulFactorLoss} is a more reliable metric than $L_p$ norms to reflect the perceptual similarity between the benign example and the adversarial example.

\begin{table}[htb]
\centering
\begin{tabular}{@{}lcccc@{}}
\toprule
Perceptual loss & $L_0$ & $L_2$ & $L_\infty$ & $MulFactorLoss$ \\
\midrule
airplane & 1 & 210 & 210 & 0.34\\
automobile & 1 & 2 & 2 & 0.01\\
bird & 2 & 397 & 203 & 2.40\\
cat & 1 & 125 & 125 & 0.65\\
deer & 1 & 69 & 69 & 0.39\\
dog & 1 & 93 & 72 & 1.01 \\
frog & 2 & 462 & 247 & 1.55\\
horse & 1 & 75 & 75 & 0.15\\
ship & 1 & 5 & 5 & 0.04\\
truck & 1 & 47 & 47 & 0.06\\
\bottomrule
\end{tabular}
\caption{Perceptual loss results of adversarial examples generated by GreedyFool over DenseNet}
\label{table3}
\end{table}

\subsection{Misclassification ratio}
According to the purpose of the tasks, adversarial attacks could be classified as non-targeted attacks and targeted attacks. Non-targeted attacks aim to mislead the DNN to classify images into any wrong classes, while targeted attacks aim to mislead the DNN to classify images into a specific target class.



We test 1,000 random images to compute the misclassification ratio for both non-targeted and targeted attacks. The results demonstrate that GreedyFool achieves 100\% success rate for both non-targeted attacks and targeted attacks.

When the perturbations are sufficient, any images could be misclassified as a specified target class by DNNs. However, adversarial attacks do not always achieve a 100\% success rate because some of them need to establish the box-constrained parameters to limit perturbations. GreedyFool utilizes greedy approximation to release this   constraint, which could automatically generate imperceptible adversarial examples in a 100\% success rate.

\subsection{Computation cost}
Computation cost refers to the running time for attackers to synthesize an adversarial example, which is employed to evaluate the attack time cost. The computation cost of GreedyFool is affected by several factors, including the feedback time of the DNN model, convergence speed of differential evolution, number of greedy approximations, running environment, etc.

The experiment chooses 30 random images from CIFAR-10 to test the actual running time of GreedyFool. On average, GreedyFool consumes approximately 38.44 seconds to carry out an adversarial attack over DenseNet on CIFAR-10, and 94.62 seconds over Inception V3 on ImageNet.

\section{Conclusion}
\label{sec:conclusion}
Adversarial attack against neural networks is a serious threat to safety-critical systems. Existing works lack sufficient consideration of the HVS, which produces noticeable artifacts in adversarial examples. In order to obtain sufficient imperceptibility, we investigate the HVS and identify four primary factors affecting the perceptibility of the human eyes: JND, Weber-Fechner law, texture masking and channel modulation.

Based on these factors, we design a pixelwise multi-factor metric to define the perceptual loss between benign images and adversarial examples. To test the multi-factor metric, we propose a black-box approach that is referred to as GreedyFool to generate adversarial examples using forward-propagation. We implement GreedyFool and conduct the adversarial attack in ImageNet on the Inception V3, CIFAR-10 on DenseNet. The experimental results show that GreedyFool has greater imperceptibility than state-of-the-art pixelwise methods, which achieves a 100\% success rate in a black-box manner.

{\small
\bibliographystyle{ieee_fullname}
\bibliography{PaperForReview}
}

\end{document}